\documentclass[numbers, sort&compress]{article}

\usepackage[preprint]{nips}

\usepackage[utf8]{inputenc} 
\usepackage[T1]{fontenc}    
\usepackage{hyperref}       
\usepackage{url}            
\usepackage{booktabs}       
\usepackage{amsfonts}       
\usepackage{nicefrac}       
\usepackage{microtype}      
\usepackage{xcolor}         

\usepackage{tikz}
\usepackage{amsmath}
\usepackage{subcaption}
\usepackage{multirow}
\usepackage{pifont}
\usepackage{listings}
\usepackage[noend, ruled, vlined]{algorithm2e}
\usepackage{subcaption}
\usepackage{multirow}
\usepackage{array}
\usepackage{graphicx}
\usepackage{color}

\definecolor{ZpfGreen}{RGB}{0,100,0}
\definecolor{ZpfRed}{RGB}{255,0,102}

\usepackage{filecontents}

\newcommand{\stitle}[1]{\vspace*{0.4em}\noindent{\bf #1.\/}}

\newcommand{\squishlist}{
	\begin{list}{$\bullet$}
		{ \setlength{\itemsep}{1pt}
			\setlength{\parsep}{1pt}
			\setlength{\topsep}{2.5pt}
			\setlength{\partopsep}{0.5pt}
			\setlength{\leftmargin}{1em}
			\setlength{\labelwidth}{1em}
			\setlength{\labelsep}{0.6em}
		}
	}
	\newcommand{\squishend}{
	\end{list}
}

\newcommand{\name}{{PSA}}

\title{Progressive Sparse Attention: Algorithm and System Co-design for Efficient Attention in LLM Serving}

\author{%
    Qihui Zhou \\
    The Chinese University of Hong Kong \\
    \texttt{qhzhou@cse.cuhk.edu.hk} \\
  \And
    Peiqi Yin \\
    The Chinese University of Hong Kong \\
    \texttt{pqyin@cse.cuhk.edu.hk} \\
  \And
    Pengfei Zuo\thanks{Pengfei Zuo is the corresponding author.} \\
    Huawei Cloud \\
    \texttt{pengfei.zuo@huawei.com} \\
  \And
    James Cheng \\
    The Chinese University of Hong Kong \\
    \texttt{jcheng@cse.cuhk.edu.hk} \\
}

\begin{document}

\maketitle

\begin{abstract}
Processing long contexts has become a critical capability for modern large language models (LLMs). However, serving long-context LLMs comes with significant inference costs due to the high memory overhead of the key-value (KV) cache.
Existing work leverages dynamic sparse attention algorithms (DSAes) to mitigate the KV cache overhead, but these algorithms rely on top-$k$ KV cache selection, which results in a trade-off between accuracy and efficiency. A larger $k$ improves accuracy but decreases efficiency, while a smaller $k$ boosts efficiency but compromises accuracy.
To overcome this trade-off, this paper presents \name, a \underline{P}rogressive \underline{S}parse \underline{A}ttention mechanism that integrates algorithmic innovations with system co-design to achieve both high inference accuracy and improved efficiency in LLM serving. The \name~algorithm adaptively adjusts the KV cache budget of different tokens and layers according to their real attention weight distributions, rather than relying on a fixed budget $k$. This enables high accuracy while minimizing KV cache usage. To further enhance execution efficiency, we introduce a pipelined iteration scheme that reduces CPU-GPU interleaving and synchronization overhead during PSA computation. Additionally,  we implement unified GPU memory management that optimizes PSA's memory utilization by accounting for uneven memory requirements across different model layers.
Extensive experimental results demonstrate that \name~reduces KV cache usage for attention computation by up to 2.4$\times$ and 8.8$\times$, and increases end-to-end serving throughput by up to 1.4$\times$ and 2.0$\times$, compared to state-of-the-art DSAes and systems without sparse attention, respectively.
\end{abstract}

\section{Introduction}

In recent years, the rapid advancement of large language models (LLMs) has reshaped our daily lives. As the demand for long-context applications, such as sophisticated reasoning~\cite{cot, tot, got} and document analysis~\cite{summ1, summ2}, continues to grow, the ability of LLMs to process long sequences has become increasingly critical. Numerous corporations and organizations have responded by developing and releasing long-context LLMs, such as DeepSeek~\cite{deepseekai2024deepseekv3technicalreport}, Gemini~\cite{gemini}, Llama~\cite{llama3.1}, and LWM~\cite{lwm}.

However, serving long-context LLMs has extremely high inference cost due to the substantial memory footprint of the key-value (KV) cache. During the processing of long sequences, these models generate the KV cache that is not static like the model weights but grows linearly with the sequence length, often surpassing the size of the model weights. For instance, the Llama-3.1 8B model, when operating with a 128K context length, requires up to 62 GB of memory to store the KV cache of a single request while the size of the model weights is only 16 GB. The giant KV cache places a significant burden on the limited GPU memory, leading to small inference batch size and low throughput. 

To reduce the inference cost of long-context LLMs, previous works~\cite{streamingllm, h2o, scissorhands, snapkv, infllm, infinigen, quest, arkvale} have explored methods to compress the KV cache required during generation. They have demonstrated that a small portion of critical tokens largely determines the generated token, and the criticalities of the tokens vary with different query tokens. Specifically, during self-attention computation, the attention scores (referred to $Q^{T}K$ in this paper) of these critical tokens are significantly larger than those of other tokens. As a result, the attention computation for each query token can be accurately approximated by involving only the KV cache of its critical tokens.

Based on the observation, dynamic sparse attention algorithms (DSAes)~\cite{infllm, quest, arkvale} for KV cache have been proposed to conduct the attention process in a select-then-compute manner. To be specific, DSAes manage the KV cache at the block level, where each block contains consecutive tokens. For each KV block, DSAes maintain small metadata derived from the token keys to represent the tokens in the block. During inference, DSAes first estimate the criticalities of all KV blocks using their metadata and the query token. They then select the top-$k$ most critical KV blocks to perform the approximate attention. The value of $k$ is predefined to balance efficiency and accuracy (e.g., 64, 128). By offloading non-critical KV blocks to host memory and loading only the selected KV blocks into GPUs, DSAes significantly reduce the GPU memory consumption, thereby enabling the efficient processing of long sequences.

Despite the theoretical potential of DSAes in reducing the computation and memory consumption of long-context LLM serving, we have found that existing DSAes face a dilemma: they tend to either produce low accuracy or achieve low KV cache reduction. The underlying issue lies in the reliance on top-$k$ selection, which assigns a uniform KV cache budget to each query token during attention computation. However, the sequence lengths of different requests are different and the attention weight ($Softmax(Q^{T}K/\sqrt{d})$) distributions of query tokens are diverse. As a result, it is hardly possible for developers to manually set a uniform budget that meets the accuracy requirements for all requests. While utilizing a larger budget can help mitigate accuracy loss, this strategy inevitably leads to low KV cache reduction, thereby sacrificing the efficiency of DSAes.

To break the dilemma, this paper proposes \name, a progressive dynamic sparse attention mechanism that integrates algorithmic innovations with system optimizations to achieve both high inference accuracy and improved performance. The core idea of \name~is the use of \emph{a threshold-based selection scheme} rather than the conventional top-$k$ selection. The threshold is defined as the minimum accumulated attention weight required for each layer of a query token, e.g., 95\%, and is configured by the LLM serving provider. Leveraging the weight threshold, \name~is able to minimize KV cache usage for each layer of a query token while retaining high accuracy across various attention weight distributions through progressive attention computation. 
Specifically, \name~performs attention at a fine-grained KV block level. For each layer of a query token, \name~first estimates the criticalities of all KV blocks within that layer. Then, starting from the most critical blocks, \name~computes partial attention results and continuously monitors the accumulated attention weights. \name~continues until the accumulated attention weights exceed the predefined threshold.

However, fine-grained attention computation at the KV block level in practical LLM serving systems is inefficient due to excessive CPU-GPU interleaving and synchronization overhead. To address this, we introduce pipelined iteration execution for PSA, which utilizes separate threads and CUDA streams to overlap KV block loading from CPU host memory with attention computation on GPUs. Additionally, a verifying GPU kernel eliminates unnecessary CPU-GPU synchronization by directly monitoring accumulated attention weights and notifying the CPU asynchronously when the predefined threshold is met. These optimizations enhance GPU utilization and improve PSA efficiency.

In addition, \name~is incompatible with the KV cache management in modern LLM serving systems, which is typically performed in a layer-wise manner. Specifically, each model layer receives equal GPU memory capacity for the KV cache and operates in isolation from other layers. However, we have observed that the attention weight skewness of different layers varies significantly, i.e., layers with low skewness access far more KV blocks during attention than those with high skewness, resulting in lower KV block cache hit ratios. To improve GPU memory utilization and the overall cache hit ratio, we propose a unified KV cache management strategy that merges the GPU memory of all layers into a unified KV block pool.

We implement \name~in vllm~\cite{vllm} and evaluate it using two representative long-context LLMs on the LongBench~\cite{longbench} dataset. Extensive experimental results demonstrate that \name~reduces the amount of accessed KV cache by up to 2.4$\times$ while maintaining the same accuracy requirement, compared to existing DSAes. In addition, \name~increases serving throughput by up to 1.4$\times$. The source code of \name~is released at Github~\footnote{\url{https://github.com/ASISys/PSAttention}}.
To summarize, this paper makes the following contributions:

\begin{itemize}

    \item We provide a comprehensive analysis of existing DSAes in long-context LLM serving and identify the limitations of the top-$k$ KV cache selection strategy.
    
    \item We introduce \name~, an efficient progressive sparse attention mechanism that addresses the limitations of existing DSAes by combining the algorithmic innovation of PSA with system optimizations, such as pipelined iteration execution and unified GPU memory management, to enhance performance.
    
    \item We implement \name~on a modern LLM serving system and demonstrate its superior performance over state-of-the-art DSAes in terms of both accuracy and efficiency.
    
\end{itemize}

\section{Background and Motivation}
In this section, we first introduce the fundamentals of generative LLM inference ($\S$~\ref{subsec:llm-basics}) and then investigate existing dynamic sparse attention algorithms ($\S$~\ref{subsec:dsaes}). Finally, we analyze the issues of the top-$k$ KV cache selection strategy employed by existing algorithms ($\S$~\ref{subsec:challenges}).

\subsection{Generative LLM Inference Basics}\label{subsec:llm-basics}
\stitle{Transformer Architecture}
The transformer has emerged as the standard model for generative LLM inference, widely adopted in popular LLMs such as GPTs~\cite{gpt4} and Llamas~\cite{llama3.1}. These LLMs are typically composed of a chain of transformer layers, each containing two modules: self-attention and feed-forward network (FFN). During inference, the input query token list $X=[x_{1},x_{2},...x_{N_{q}}]$ for each layer is first multiplied by three weight matrices $W_{q}$, $W_{k}$, and $W_{v}$ to generate query ($Q \in R^{N_{q} \times d}$), key ($K \in R^{N_{kv} \times d}$), and value ($V \in R^{N_{kv} \times d}$) matrices, where $N_{q}$ is the number of query tokens, $N_{kv}$ is the number of KVs, and $d$ is the hidden dimension. The self-attention is then conducted using the $Q$, $K$, and $V$ as follows:
\begin{equation}
        S = \frac{Q \cdot K^{T}}{\sqrt{d}},\quad
        P = softmax(S),\quad
        O = P \cdot V
        \notag
\end{equation}
Here, $P$ represents the \emph{attention weights}, with $P_{i, j}$ indicating the \emph{importance} of $K_j$ to query token $i$. The attention output $O$ is then fed into the FFN module, which comprises two consecutive linear projections with a non-linear activation operation between them. The output of the FFN serves as the input of the next transformer layer.

\stitle{Auto-regressive Generation}
In generative LLM inference, the processing of each request involves two key phases: \textit{the prefill phase} and \textit{the decoding phase}. The prefill phase processes the tokens of an input prompt in parallel and produces the first output token, whose latency is called time-to-first-token (TTFT). Taking the first output token as input, the LLMs generate the next new token during the decoding phase. This newly generated token then serves as the input token for the next iteration, resulting in an auto-regressive process for token generation. This generation process continues until the special stopping token \textit{EOS} is generated or the number of output tokens reaches a predefined maximum limit. The latency taken to generate each output token in the decoding phase is called time-between-token (TBT). 

\stitle{KV Cache}
During the generation of new tokens, the key and value vectors of all previous tokens are required for the self-attention computation. As a result, they are generally cached in GPU memory to avoid repeated computation, which are referred to as KV cache. Since the size of the KV cache grows with the execution of decoding, existing LLM serving systems generally employ PagedAttention~\cite{vllm} to divide GPU memory into fixed-size blocks and store the KV cache in multiple discontinuous memory blocks to avoid memory fragmentation. 


\subsection{Dynamic Sparse Attention}\label{subsec:dsaes}
\begin{figure}[!t]
	\centering
	\includegraphics[width=0.8\columnwidth]{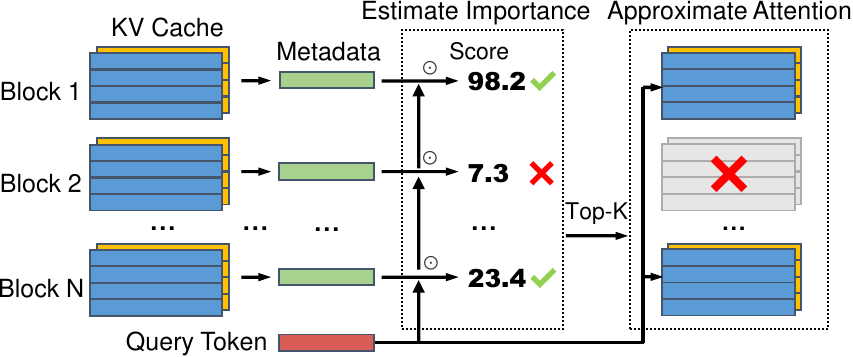}
	\caption{The workflow of the dynamic selection algorithm.}
	\label{fig:bk:dsaes}
\end{figure}

\stitle{KV Cache Eviction} Due to the auto-regressive nature of LLMs, generating each token necessitates loading the entire KV cache from GPU HBM to on-chip SRAM. For long-context LLM serving, the large size of the KV cache results in significant time and space overheads. Despite the large size of the KV cache, recent works~\cite{h2o, streamingllm} have observed that the attention computation of LLM is highly sparse, with only a small portion of tokens contributing to the majority of attention weights. Based on this observation, several KV cache eviction algorithms~\cite{streamingllm, h2o, snapkv, fastgen, scissorhands} have been proposed which statically discard the KV cache of unimportant tokens to maintain a smaller KV cache size. However, these eviction methods determine which tokens to discard based on historical information or current query tokens, but the discarded tokens might be important for future tokens~\cite{quest}, potentially leading to the loss of important information.

\stitle{KV Cache Selection} To prevent information loss, dynamic sparse attention algorithms (DSAes)~\cite{arkvale, infllm, quest} have been proposed. Figure~\ref{fig:bk:dsaes} illustrates the general workflow of existing DSAes. Instead of permanently discarding unimportant tokens, they retain all KV cache and dynamically select a small portion of the critical KV cache for each query token for attention computation. Since not all KV cache is required for attention computation, DSAes allow the KV cache to be offloaded to host memory and load only the selected KV blocks into GPUs each time to reduce GPU memory consumption. To speed up the selection process, inspired by the block-level memory allocation of the KV cache in PagedAttention~\cite{vllm}, DSAes divide the KV cache into blocks and select the KV cache at the block level. For each KV block, DSAes construct metadata vectors to represent the tokens within it. Different DSAes propose various metadata construction methods, ranging from simply calculating the mean values of the token keys to finding the bounding cuboid of the token keys. Regardless of the methods, the size of metadata is much smaller than the KV block. To estimate the importance of KV blocks to query tokens, DSAes compute dot products between the metadata vectors and the query tokens to obtain approximate attention scores for all blocks. DSAes then select the top-$k$ most critical KV blocks to perform the approximate attention. 

We notice that there are recent works~\cite{infinigen, tokenselect} that propose to select KV cache at the granularity of token instead of blocks. Although such token-level selection can identify important tokens more accurately in theory, it is overly granular and incurs significant runtime overhead. Firstly, the identification of important tokens is costly in terms of both memory and computation. Existing token-level DSAes require the keys of all previous tokens to compute attention scores and then identify the important tokens, making the selection process slow when the context is lengthy. Secondly, token-level selection leads to significant overhead in GPU cache management due to the increased number of cache slots required for KV cache eviction and loading. In contrast, block-level selection achieves a balance between accuracy and performance overhead, as a result of which, we focus on block-level DSAs in this work.



\subsection{The Dilemma of the Top-$k$ Selection}\label{subsec:challenges}

\begin{figure}[t]
	\centering
	\begin{subfigure}[b]{0.35\columnwidth}
		\centering
		\includegraphics[width=\textwidth]{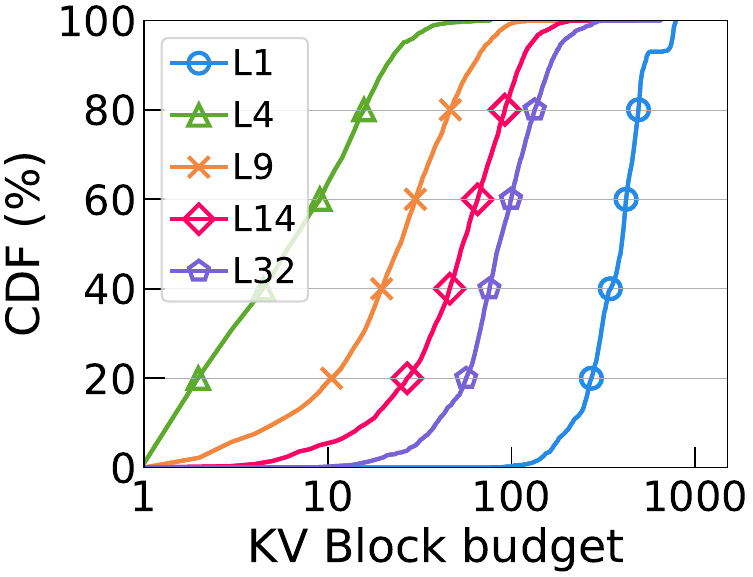}
		\caption{QMSum}\label{fig:bk:sparsity-variation-qasper}
	\end{subfigure}
	\begin{subfigure}[b]{0.35\columnwidth}
		\centering
		\includegraphics[width=\textwidth]{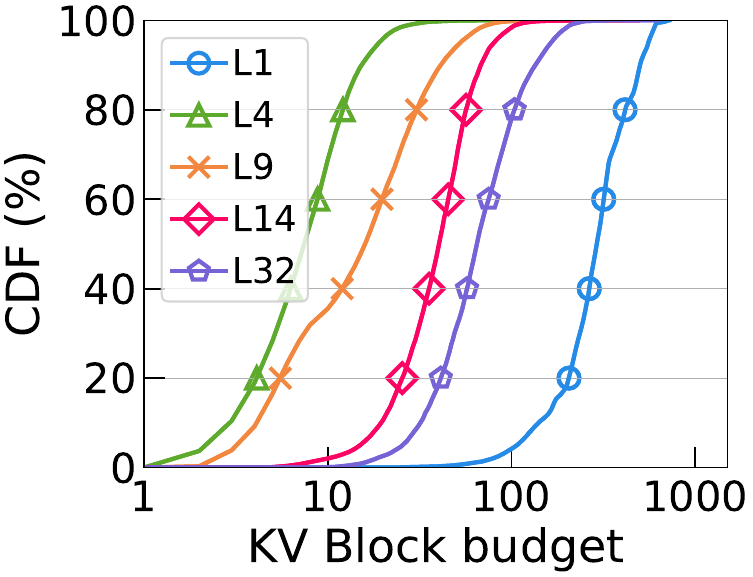}
		\caption{GovReport}\label{fig:bk:sparsity-variation-govreport}
	\end{subfigure}
	\caption{The CDF of the numbers of query tokens that achieve 0.95 out of 1 total attention weights under varying block budgets. \textit{(The block size is set to 32. The model used is LWM-Text-7B~\cite{lwm}.)}}
	\label{fig:bk:sparsity-variation}
\end{figure}

Existing DSAes rely on top-$k$ selection, where the same KV cache budget (i.e., the $k$ value) is applied across all layers and query tokens during attention computation. To achieve high inference accuracy, the top-$k$ KV blocks selected by DSAes should have a sufficiently high cumulative attention weight with the query token (e.g., 95\%). However, we observe that the fixed top-$k$ selection fails to consistently achieve this due to variability in sparsity across tokens and layers. 

\stitle{1) Attention sparsity varies across tokens} We evaluate the sparsity variation using two popular datasets QMSum~\cite{qmsum} and GovReport~\cite{govreport} from LongBench~\cite{longbench}, as shown in Figure~\ref{fig:bk:sparsity-variation}. It is observed that the number of KV blocks needed to achieve 95\% accumulated attention weight varies significantly across query tokens. For example, in Layer 32 (L32) of GovReport, 20\% of query tokens require fewer than 50 KV blocks, 60\% need between 50-100 KV blocks, and the remaining 20\% need more than 100 KV blocks. 

\stitle{2) Attention sparsity also varies across layers} We also observe that the number of KV blocks needed to achieve the same accumulated attention weight varies significantly across layers. For example, in Layer 9 (L9) of QMSum, 80\% of query tokens require fewer than 50 KV blocks, while in Layer 1 (L14), the number decreases to 40\%. 

As a result, it is hardly possible to set a uniform KV block budget that can satisfy the accuracy requirements of all tokens. A small $k$ value may lead to insufficient attention weight for most tokens, resulting in low accuracy, while a large $k$ value risks over-selecting KV blocks, harming performance.

\section{The \name~Methodology}
In this section, we present \name, a progressive sparse attention mechanism that integrates algorithmic innovations with system optimizations to achieve both high inference accuracy and improved performance. We first show a high-level system overview for \name~($\S$~\ref{subsec:overview}) and then detail the key algorithm innovations and system optimizations proposed in \name, including progressive attention kernel ($\S$~\ref{subsec:progressiveattention}) and unified memory management ($\S$~\ref{subsec:unifiedmemory}).

\subsection{System Overview}\label{subsec:overview}

\begin{figure}[!t]
	\centering
	\includegraphics[width=0.7\columnwidth]{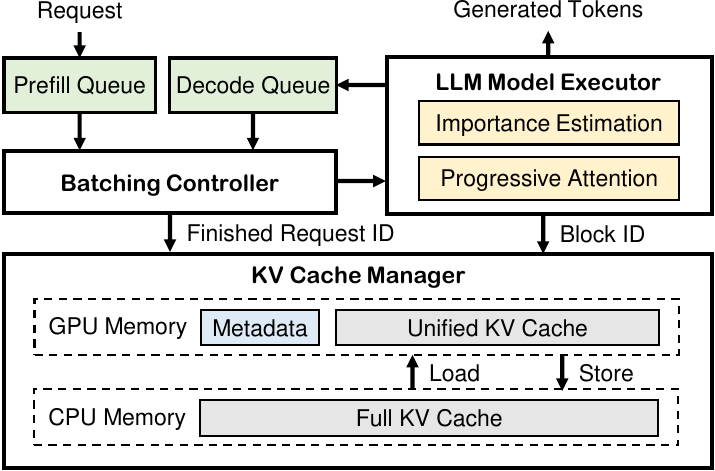}
	\caption{The system architecture for \name.}
	\label{fig:design:architecture}
\end{figure}

Figure~\ref{fig:design:architecture} illustrates the overall system architecture for \name, which comprises three key components: the batch scheduler, the model executor, and the KV cache manager.

\squishlist
\item \textit{The Batch Controller} is responsible for grouping incoming requests into batches using dynamic batching techniques~\cite{orca} in a first-come-first-serve (FCFS) manner. These batches are subsequently forwarded to the model executor for processing. The batch controller ensures that the KV cache blocks required by the requests in the batch for attention computation can fit in GPU memory. Once a request is finished, the batch controller sends the request ID to the KV cache manager, signaling it to release the corresponding KV blocks.

\item \textit{The Model Executor} receives batches from the batch controller and performs the computation of model forwards. The model executor replaces the standard attention computation with the progressive attention mechanism ($\S$~\ref{subsec:progressiveattention}) for decoding requests. During batch execution, the model executor transmits the newly generated KV cache to the KV cache manager and fetches the KV block \textit{metadata} of the requests from the KV cache manager. Additionally, it sends the IDs of the KV blocks required for the attention computation to the KV cache manager to trigger the data loading. The \textit{metadata} of KV blocks is used to estimate the criticality of KV blocks for a query token. By default, \name~utilizes the cuboid-mean method~\cite{arkvale} to construct the metadata for KV blocks due to its high recall accuracy. However, other metadata construction methods~\cite{infllm, quest} can be easily integrated into \name.

\item \textit{The KV Cache Manager} maintains a hierarchical KV block storage between GPU memory and host memory. Both GPU and host memory are organized into blocks to mitigate memory fragmentation~\cite{vllm}. The KV cache manager receives newly generated KV cache data from the model executor and writes them to the corresponding KV blocks in GPU memory. Once a KV block reaches its capacity, it is flushed to host memory asynchronously, and its associated metadata is created. The metadata is retained in GPU memory due to its small size and is utilized in every attention computation. The remaining GPU memory is used to cache frequently accessed KV blocks ($\S$~\ref{subsec:unifiedmemory}). 

\squishend

\subsection{Progressive Sparse Attention Algorithm}\label{subsec:progressiveattention}
\begin{algorithm}[t]
	\caption{The PSA algorithm.}
	\label{alg:progressattention}
	\DontPrintSemicolon
	\SetAlgoLined
	\LinesNumbered
	\SetNoFillComment
	\SetKwComment{Comment}{$\triangleright$\ }{}
	\BlankLine
	
	\KwIn{query vector $q$ and KV block indices $\mathcal{B}$.}
	\KwOut{approximate attention output $O_{acc}$.}
	
	\BlankLine

	\SetKwFunction{FMain}{PSAttention}
	\SetKwProg{Fn}{def}{:}{}
	\Fn{\FMain{q, $\mathcal{B}$}}{
        \textit{$O_{acc}$} $\leftarrow$ $[0,\ ...\ ,\ 0]_{d}$ \tcp*[f]{Init output} \\
        $AS_{acc}$ $\leftarrow$ $0$ \\
        $AS_{min}$ $\leftarrow$ $MAX\_VAL$ \\
        $CS$ $\leftarrow$ $q \cdot \mathcal{B}_{meta}$  \tcp*[f]{Criticality scores} \\
        $\mathcal{B}'$ $\leftarrow$ Rank $\mathcal{B}$ by $CS$ \\
        $N_{left}$ $\leftarrow$ \texttt{Len($\mathcal{B}$)} \tcp*[f]{Num of blocks left} \\
		\ForEach{\textnormal{block} $b$ \textnormal{in} $\mathcal{B}'$}{
            $KV_{b}$ $\leftarrow$ \texttt{Load($b$)} \tcp*[f]{KV cache of $b$} \\
            $O$, $AS$ $\leftarrow$ \texttt{Attention($q$, $KV_{b}$)} \\
            $O_{acc}$ $\leftarrow$ $(AS_{acc} \cdot O_{acc} + AS \cdot O)\ /\ (AS_{acc} + AS)$ \\
            $AS_{acc}$ $\mathrel{{+}{=}}$ $AS$ \\
            $AS_{min}$ $\leftarrow$ \texttt{MIN($AS$, $AS_{min}$)} \\
            $N_{left}$ $\mathrel{{-}{=}}$ $1$  \\
            $AS_{total}$ $\mathrel{{=}}$  $(AS_{acc} + AS_{min} \cdot N_{left})$ \\
            $\mathcal{P}_{acc}$ $\leftarrow$ $AS_{acc}$\ /\ $AS_{total}$ \\
            \If{ $\mathcal{P}_{acc} >$ \textnormal{attention weight threshold $\epsilon$}}{
                \textbf{break} \\
            }
		}
		\KwRet \textit{$O_{acc}$}
	}
\end{algorithm}

\name~leverages a threshold-based KV block selection algorithm rather than the top-k-based selection, which adaptively assigns the KV block budget for each layer of each query token according to its attention weight distributions. Specifically, \name~executes the attention computation progressively at the KV block level and keeps monitoring the accumulated attention weight of the computed KV blocks. The attention computation of a query token is immediately terminated once the accumulated attention weight reaches a given threshold (e.g., 95\%), thereby avoiding unnecessary KV cache loading and extra computation. 

Algorithm~\ref{alg:progressattention} outlines the PSA algorithm. Given the query vector and the KV block indices of a decoding request, \name~first fetches the metadata of these KV blocks, $\mathcal{B}_{meta}$, and uses it to calculate the criticality score of each KV block, $CS$, for the query ($Line~5$), similar to existing DSAes. Then, instead of selecting the top-k KV blocks with the highest criticality scores for attention, \name~ranks these KV blocks according to their criticality scores ($Line~6$) and begins attention computation from the most critical KV blocks to the least critical KV blocks ($Lines~8-10$). 

\name~\textit{progressively} executes the attention computation at the granularity of a microbatch of KV blocks, where a microbatch consists of $m$ KV blocks—$m$ being the minimum number of KV blocks required for a single attention computation, designed to reduce the frequency of attention kernel calls. In the example shown in Algorithm~\ref{alg:progressattention}, $m$ is set to 1.  After each attention computation, a partial output, $O$, and the sum of the token attention scores, $AS$, are returned ($Line~10$). \name~then merges the $O$ into the accumulated output, $O_{acc}$ ($Line~11$), and updates the accumulated sum of the exponentiation of attention scores, $AS_{acc}$ ($Line~12$). 

\name~then computes the ratio of the accumulated attention score, $AS_{acc}$, to the total attention score, $AS_{total}$. However, $AS_{acc}$ is unknown before performing attention computation on all KV blocks. To address the issue,
\name~records the minimum block-level attention score encountered so far, $AS_{min}$ ($Line~13$), and uses it as the upper-bound attention score of the remaining blocks that have not been processed to estimate $AS_{total}$ ($Line~15$), calculated as $AS_{acc} + AS_{min} * N\_left$. Finally, the accumulated attention weights, $P_{acc}$, can be estimated using $AS_{acc}$ and  $AS_{total}$ ($Line~16$). When $P_{acc}$ exceeds the threshold, $O_{acc}$ is returned and the attention computation completes ($Lines~17-18$).

\begin{figure}[!t]
	\centering
	\includegraphics[width=0.75\columnwidth]{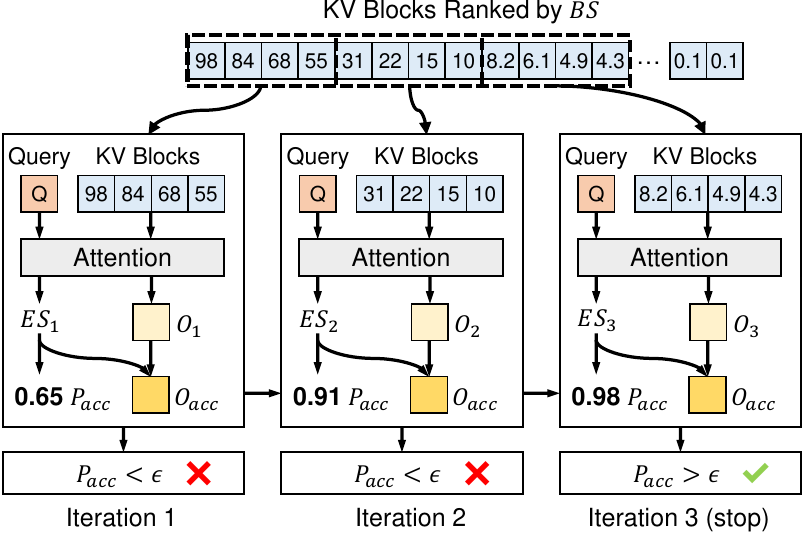}
	\caption{An example of executing the \name~algorithm.}
	\label{fig:design:progressive-attention}
\end{figure}

Figure~\ref{fig:design:progressive-attention} illustrates an example of executing the \name~ algorithm. Given a list of KV blocks ranked by criticality scores and a query Q, \name~first computes the attention weights between Q and the first microbatch of KV blocks (98, 84, 68, 55), resulting in an accumulated attention weight of $P_{acc} = 0.61$. Since 0.61 is smaller than the threshold $\epsilon$ (= 0.98), \name~proceeds to compute the attention weights between Q and the second microbatch of KV blocks (31, 22, 15, 10), yielding $P_{acc} = 0.91$. As 0.91 is still below $\epsilon$, \name~continues with the third microbatch of KV blocks (8.2, 6.1, 4.9, 4.3), eventually reaching an accumulated attention weight of $P_{acc} = 0.98$. When $P_{acc}$ equals $\epsilon$, the attention computation completes.

Moreover, Algorithm~\ref{alg:progressattention} is presented for a single decoding request for clarity but be easily extended to support batched decoding requests. To execute a batch of multiple query tokens, each corresponding to a decoding request, \name~is invoked with all query tokens synchronously. Since each query token in the batch may require a different number of KV blocks to reach the attention weight threshold, we examine the current accumulated attention weights for all query tokens at the end of each iteration. Query tokens whose attention weights have already surpassed the threshold are removed from further processing. The batched \name~process continues until no query token remains in the batch.

\subsection{Pipelined Iteration Execution}\label{subsec:execution}
Nevertheless, implementing the PSA algorithm in practical LLM serving systems presents challenges for achieving efficient execution. The \name~algorithm breaks the attention process into multiple iterations. In each iteration, the algorithm first prepares the key-value (KV) blocks by loading any missing blocks from host memory and updating the GPU HBM cache. It then performs the standard attention computation on the prepared KV blocks and subsequently updates the accumulated attention weights while checking whether the predefined threshold is met. This dynamic multi-iteration computation pattern can lead to two performance issues: (1) low GPU utilization due to the interleaving of CPU data preparation and GPU computation, and (2) significant synchronization overhead between CPUs and GPUs due to the frequent transfer of attention weights. 

To address these issues, we present pipelined iteration execution for PSA. Specifically, to enhance GPU utilization, we assign the data preparation and attention computation to separate threads, ensuring that data loading and computation can proceed concurrently without blocking each other. To further maximize parallelism, we utilize separate CUDA streams for these operations, allowing data transfers and kernel executions to overlap in a pipelined manner. This design minimizes idle GPU time and improves overall execution efficiency by ensuring that the GPU is continuously utilized for computation while data preparation occurs in parallel.

To mitigate synchronization overhead, we design a verifying GPU kernel that updates and checks the accumulated attention weights directly on GPUs,  eliminating the need to transfer these weights. Once the accumulated attention weights reach the predefined threshold, the verifying kernel writes a signal variable allocated in pinned host memory using the zero-copy technique, notifying the CPUs to terminate the attention process.

\subsection{Unified Memory Management}\label{subsec:unifiedmemory}

GPU memory management in existing LLM serving systems, such as vllm~\cite{vllm}, is typically conducted in a layer-wise manner. Specifically, the available GPU memory capacity is first determined through a profiling execution. Then each layer is allocated with a GPU tensor with an equal capacity of GPU memory and operates in isolation from the other layers. This design choice is based on the fact that the transformer layers of modern LLMs are identical in terms of memory usage patterns. To be specific, when executing an iteration for a batch of requests, each request accesses the same amount of KV blocks across all layers. 

However, directly applying such a layer-separated memory management approach in ~\name~leads to low GPU memory utilization. The reason is that ~\name~adaptively adjusts the KV blocks involved in the attention computation according to the attention weight distributions through progressive attention. As shown in Figure~\ref{fig:bk:sparsity-variation}, the attention weight skewness of different layers varies significantly. The layers with low skewness access far more KV blocks during attention than those with high skewness, leading to a lower cache hit ratio. 

To address this problem, ~\name~proposes a unified GPU memory management strategy. Specifically, ~\name~first determines the available GPU memory capacity through a profiling execution like existing systems, and allocates a single GPU tensor of the capacity. The GPU tensor is then divided into equal-sized slots to store KV blocks, which are managed by the KV cache manager. During runtime, all operations on the KV blocks across layers including allocation, deallocation, and loading are sent to the KV cache manager for processing. 

Currently, we employ the least recently used (LRU) policy as the KV cache eviction policy, which leverages the semantic similarity between consecutive query tokens during decoding, as reported in prior works~\cite{infllm, tokenselect}. Specifically, the query vectors of consecutive tokens generated during decoding exhibit a high similarity, leading to the selection of similar KV blocks. It is important to note that other cache eviction policies, such as the first in first out (FIFO) policy, can be easily integrated into \name, and we leave designing more sophisticated policies for future work.

\section{Implementation}\label{sec:implementation}
\stitle{Request Scheduling}
We implement \name~in vllm\cite{vllm} with about 6,000 lines of code. In vllm, a request can only be scheduled if all its KV cache can fit into GPU memory. However, with DSAes, it is not necessary for a request to have all its KV cache available for execution. In \name, a request can be scheduled as long as the required KV blocks for executing a single iteration of progressive attention can be accommodated within the GPU memory.

\stitle{Compatibility with FlashAttention~\cite{flash-attention}}
FlashAttention is an attention backend widely used in modern LLM serving system to accelerate attention computation. It avoids writing attention weights to HBM through kernel fusion, thus significantly reducing data movement between on-chip memory and HBM. However, our proposed \name~mechanism requires the storage of attention weights to estimate the accumulated attention weights. To mitigate this issue, we aggregate the attention weights of tokens within each block using on-chip memory and subsequently write the aggregated results back to HBM.

\section{Performance Evaluation}
\subsection{Experimental Setup}
\stitle{Testbed} 
Our experiments are conducted on a machine hosting an Nvidia A100 GPUs with 40 GB HBM, an AMD EPYC 7J13 CPU, and 128 GB DRAM. The GPU is connected to the host via PCIe Gen 4, providing a bandwidth of 32 GB per second.

\stitle{Models}
The experiments evaluate two widely used long-context LLM models: LWM-Text-7B~\cite{lwm} with a 1M context window and Llama-3.1-8B~\cite{llama3.1} with a 128K context window. In particular, the LWM-Text-7B model employs the same model architecture as Llama-2-7B~\cite{llama2}. These two models cover two mainstream attention methods, namely, multi-head attention (MHA) and grouped query attention (GQA).

\stitle{Workload}
We conduct experiments using several datasets from LongBench~\cite{longbench}, including HotpotQA~\cite{hotpotqa}, 2WikiMultihopQA~\cite{2wikimultihopqa}, MultifieldQA~\cite{longbench}, Qasper~\cite{qasper}, GovReport~\cite{govreport}, QMSum~\cite{qmsum}, MultiNews~\cite{multinews}, and SAMSum~\cite{samsum}. Accuracy is evaluated separately for each dataset, while efficiency is evaluated by combining the requests from all datasets into one trace. This reflects real-world LLM serving scenarios where requests from different tasks are handled simultaneously. As there is no public request arrival timestamps available in these datasets, we generate request arrival times based on a Poisson distribution with varying arrival rates, as done in prior works~\cite{orca, sarathiserve}. 

\stitle{Baselines}
We compare \name~with the following baselines: vllm~\cite{vllm}, vllm-sparse, and InfiniGen~\cite{infinigen}. vllm is a state-of-the-art LLM serving system that employs full KV cache attention without sparsity. Since no existing DSA implementations support dynamic request batching~\cite{orca}, we implement Arkvale~\cite{arkvale}, which is the state-of-the-art block-level DSA, within the vllm, creating an additional baseline termed vllm-sparse. InfiniGen~\cite{infinigen} is recognized as the state-of-the-art token-level DSA and is implemented within the offloading-based LLM inference system FlexGen~\cite{flexgen}. 

\begin{figure*}[!t]
	\centering
	\includegraphics[width=0.95\columnwidth]{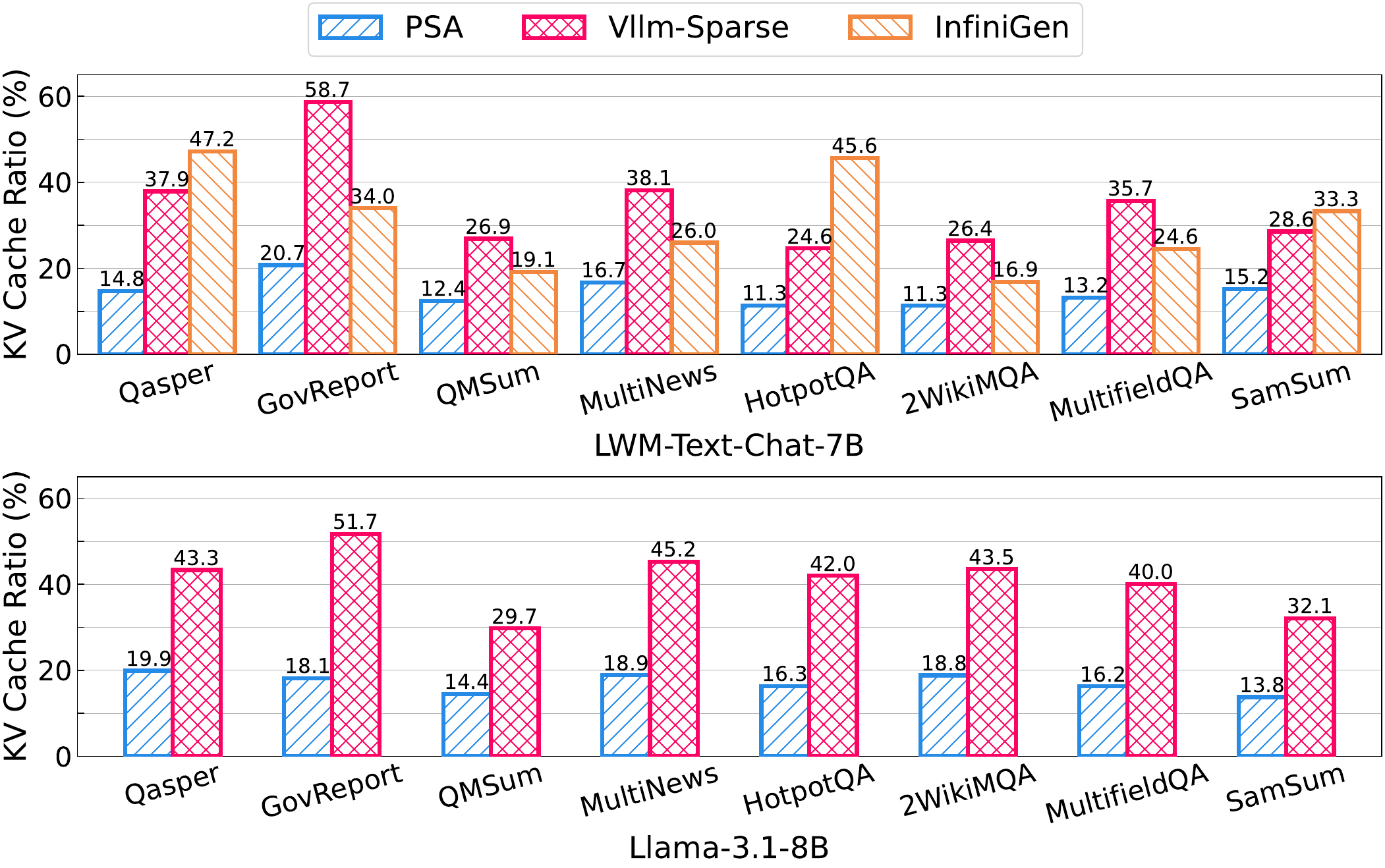}
	\caption{The average amount of KV cache required by vllm-sparse, InfiniGen, and \name~to achieve over 98\% average request accuracy when compared with vllm across different datasets from LongBench.}
	\label{fig:eval:accuracy}
\end{figure*}

\subsection{Main Results}
\stitle{KV cache reduction}
We first compare the effectiveness of vllm-sparse, InfiniGen, and \name~in reducing the KV cache used during attention computation. We ensure that the average request accuracy meets a target of $98\%$, which is typical for practical production scenarios~\cite{AdaptiveNN, Apparate}. Figure~\ref{fig:eval:accuracy} shows the average KV cache utilized by these three systems during attention computation compared with vllm using full KV cache attention. For vllm-sparse, the amount of KV cache is adjusted by varying the top-$k$ value. For \name, it is modified by setting different thresholds for the accumulated attention weights. For InfiniGen, the token budget is adjusted by varying the minimum attention score of the selected tokens. Since InfiniGen only supports MHA and does not accommodate GQA, we report its performance only on LWM-Text-7B.

As shown in Figure~\ref{fig:eval:accuracy}, \name~consistently outperforms both vllm-sparse and InfiniGen by a large margin. Specifically, \name~reduces the KV cache ratio for attention by 2.1$\times$ and 2.4$\times$ on average for LWM-Text-7B and Llama-3.1-8B, respectively, compared to vllm-sparse. This is because vllm-sparse assigns a uniform KV cache budget across all requests, leading to the over-selection of KV blocks for requests with shorter lengths and skewed attention weight distributions. In contrast, \name~dynamically adjusts KV cache budgets according to the attention weight distributions of requests, minimizing the KV cache usage. Compared to InfiniGen, \name~reduces the KV cache ratio for attention by 1.8x on average for LWM-Text-7B. This result may seem counterintuitive, given InfiniGen uses fine-grained token-level selection. However, InfiniGen uses a uniform minimal attention score threshold to selects critical tokens, which fails to account for the varying ranges of attention scores across requests. Moreover, InfiniGen adopts weight matrices compression and inter-layer prefetching to speed up the KV cache selection and loading process, but this inadvertently results in accuracy degradation in attention score estimation.

\begin{figure}[!t]
	\centering
	\includegraphics[width=0.7\columnwidth]{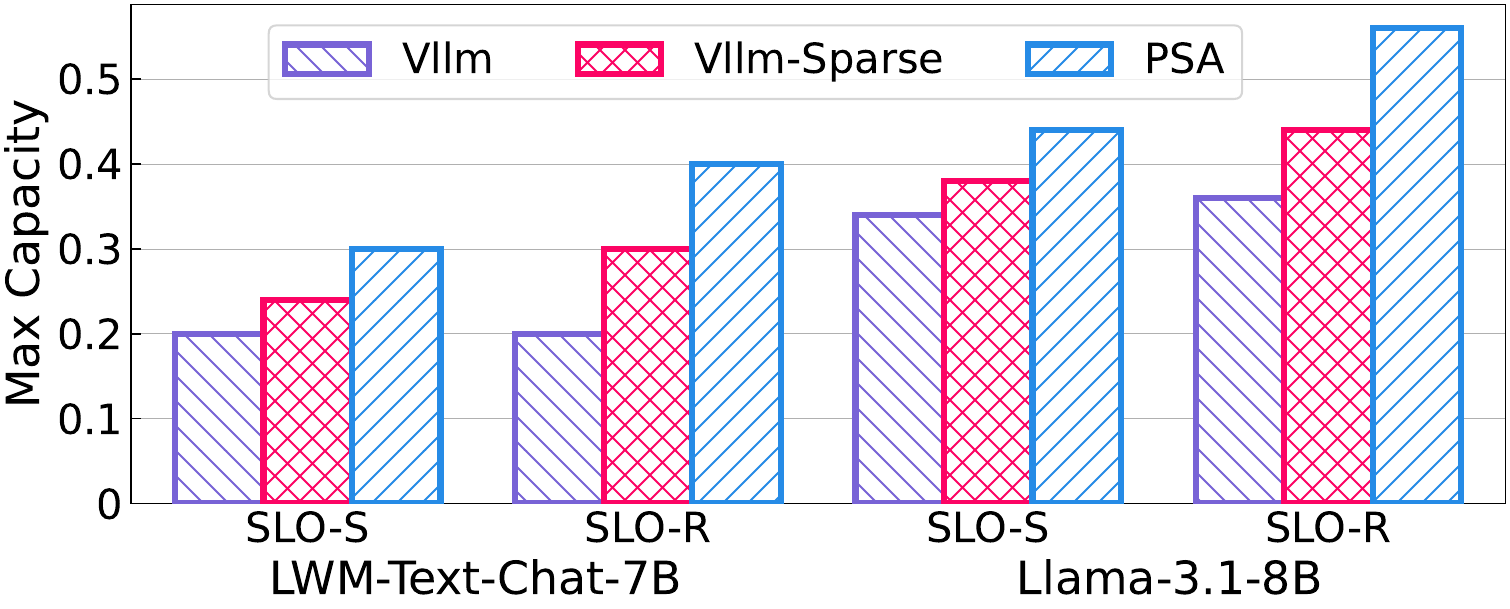}
	\caption{Maximum capacities in QPS of vllm, vllm-sparse, and \name~across each dataset under SLO constrains.}
	\label{fig:eval:goodput}
\end{figure}

\stitle{Capacity evaluation}
We measure the request throughput of vllm, vllm-sparse, and \name~under the same Service Level Objective (SLO) requirements, as shown in Figure~\ref{fig:eval:goodput}. Following prior work~\cite{splitwise, sarathiserve}, we define the strict and relaxed SLOs for the P99 TBT latency as 5$\times$ and 25$\times$ the execution time of a decoding iteration, respectively. Specifically, for the LWM-Text-7B model, the strict and relaxed SLO requirements are set to 150 and 750 milliseconds, respectively. For the Llama-3.1-8B model, the corresponding strict and relaxed SLO requirements are 100 and 500 milliseconds, respectively. In all load experiments, we ensure the sustainability of the maximum load by imposing a threshold on request queuing delay. The median scheduling delay for requests is limited to 2 seconds to prevent excessive queuing, as suggested in~\cite{sarathiserve}.

Compared to vllm, \name~achieves 1.5$\times$ and 1.3$\times$ higher throughput for the LWM-Text-7B and Llama-3.1-8B, respectively, under strict SLO requirements. With relaxed SLO requirements, \name~further increases the throughput by up to 2.0$\times$ for LWM-Text-7B and 1.5$\times$ for Llama-3.1-8B. This is because the throughput of vllm is limited by its small inference batch size, which is constrained by GPU memory. In contrast, \name~effectively reduces the amount of KV cache involved in attention computation, enabling larger inference batch sizes and higher throughput. Compared to vllm-sparse, \name~sustains up to 1.3$\times$ and 1.2$\times$ higher loads for LWM-Text-7B and Llama-3.1-8B, respectively, under strict SLO requirements. With relaxed SLO requirements, \name~sustains loads up to 1.4$\times$ and 1.3$\times$ higher loads for LWM-Text-7B and Llama-3.1-8B, respectively. This is because \name~reduces more KV cache usage compared to vllm-sparse thanks to the progressive sparse attention mechanism.



\section{Related Work}\label{sec:related-work}
\stitle{Static Sparse Attention}
Several KV cache eviction algorithms, e.g., H2O~\cite{h2o}, StreamingLLM~\cite{streamingllm}, SnapKV~\cite{snapkv}, FastGen~\cite{fastgen}, and Scissorhands~\cite{scissorhands}, have been proposed to retain only the KV cache of important tokens while discarding others to save GPU memory. However, since the importance of tokens changes during the decoding process, discarded tokens may become crucial for further computation~\cite{quest}, resulting in potential accuracy loss. 

\stitle{Dynamic Sparse Attention} DSAes, such as ArkVale~\cite{arkvale}, InfLLM~\cite{infllm}, Quest~\cite{quest}, mitigate the issue of static sparse attention by dynamically selecting a small portion of the critical KV cache for attention computation for each query token while retaining all KV cache. While existing DSAes predominantly focus on enhancing the accuracy of important KV cache identification, \name~is the first work to consider their deployment efficiency in practical LLM serving systems, achieving both high accuracy and efficiency through algorithm and system co-design.

\stitle{Token-level Sparse Attention} Recent works, such as InfiniGen~\cite{infinigen}, TokenSelect~\cite{tokenselect}, RetrievalAttention~\cite{retrievalattention}, and MagicPig~\cite{magicpig}, perform KV cache selection at the granularity of tokens. However, as discussed in \S~\ref{subsec:dsaes}, token-level selection incurs significant runtime overhead, making block-level selection a better balance between accuracy and performance overhead. Consequently, this paper focuses on block-level DSA approaches.


\stitle{Sparse Attention for Prefill and Training} There are also works~\cite{Gems, mInference, SeerAttention} that apply sparse attention to accelerate the prefill phase and training of LLMs. GemFilter~\cite{Gems} prunes unimportant tokens using attention matrices from early layers to reduce the computational load in subsequent layers. Minference~\cite{mInference} recognizes three general patterns of sparse attention in long-context LLMs and provides optimized CUDA kernels for each pattern. SeerAttention~\cite{SeerAttention} extends Minference by replacing fixed patterns with a learnable approach. Native sparse attention (NSA)~\cite{Yuan2025NativeSA} introduced by DeepSeek is the first to perform sparse attention during training.
These methods are orthogonal to the \name~proposed in this paper and can be combined with it to further enhance the overall efficiency of end-to-end LLM processing.

\stitle{Inference Parameter Offloading}
DeepSpeed Inference~\cite{deepspeed-inference, ZeRO-Infinity} offloads model parameters to host memory and fetch them on demand. Lina~\cite{lina} leverages sparse activation in mixture-of-expert (MoE) models to offload cold experts to host memory. PowerInfer~\cite{PowerInfer} utilizes the sparsity in FFN computation to offload inactive weights to host memory, thereby saving memory and computational resources. FlexGen~\cite{flexgen} offloads both model parameters and KV cache to host memory,  targeting offline processing. In contrast, \name~exploits KV cache offloading in online LLM serving by utilizing the sparsity in KV cache.


\section{Conclusion}
This paper presents \name~, an efficient progressive sparse attention mechanism to address the limitations of the top-$k$ KV cache selection in existing DSAes. By combining the algorithmic innovation of PSA with system-level optimizations, such as pipelined iteration execution and unified GPU memory management, \name~achieves both high accuracy and efficiency. Extensive experimental results demonstrate that \name~reduces KV cache usage for attention computation by up to 2.4$\times$ while maintaining the same accuracy, and increases the serving throughput by up to 1.4$\times$.



\bibliography{sample}
\bibliographystyle{plainnat}

\end{document}